\documentclass[conference]{IEEEtran}
\IEEEoverridecommandlockouts
\usepackage[T1]{fontenc}
\usepackage{cite}  
\usepackage[caption=false,font=footnotesize]{subfig}
\usepackage{amsmath,amssymb,amsfonts}
\usepackage{algorithmic}
\usepackage{cleveref}
\usepackage{graphicx}
\usepackage[hyphens]{url}
\usepackage{tikz}
\newcommand\copyrighttext{%
  \footnotesize \textcopyright 2026 IEEE. Personal use of this material is permitted.
  Permission from IEEE must be obtained for all other uses, in any current or future media, including reprinting/republishing this material for advertising or promotional purposes, creating new collective works, for resale or redistribution to servers or lists, or reuse of any copyrighted component of this work in other works.}
\newcommand\copyrightnotice{%
\begin{tikzpicture}[remember picture,overlay]
\node[anchor=south,yshift=10pt] at (current page.south) {\fbox{\parbox{\dimexpr\textwidth-\fboxsep-\fboxrule\relax}{\copyrighttext}}};
\end{tikzpicture}%
}
\usepackage{textcomp}
\usepackage{etoolbox}
\apptocmd{\thebibliography}{\setlength{\itemsep}{0pt}}{}{}
\usepackage{enumitem}
\setlist{noitemsep, topsep=0pt}
\usepackage{xcolor}
\def\BibTeX{{\rm B\kern-.05em{\sc i\kern-.025em b}\kern-.08em
    T\kern-.1667em\lower.7ex\hbox{E}\kern-.125emX}}

\makeatletter
\@ifundefined{IEEEbibfont}{}{}
\makeatother
\patchcmd{\thebibliography}{\relax}{\addtolength{\itemsep}{-2pt}}{}{}
\begin{document}

\title{Animating Petascale Time-varying Data on Commodity Hardware with LLM-assisted Scripting\\
}


\author{
  \IEEEauthorblockN{Ishrat Jahan Eliza\IEEEauthorrefmark{1}\thanks{Corresponding author: Ishrat Jahan Eliza (\protect\url{ishratjahan.eliza@utah.edu})}, Xuan Huang\IEEEauthorrefmark{1}, Aashish Panta\IEEEauthorrefmark{1}, Alper Sahistan\IEEEauthorrefmark{1},}
  \IEEEauthorblockN{Zhimin Li\IEEEauthorrefmark{2}, Amy A. Gooch\IEEEauthorrefmark{3}, Valerio Pascucci\IEEEauthorrefmark{1}}
    \IEEEauthorblockA{\IEEEauthorrefmark{1}University of Utah \quad \IEEEauthorrefmark{2}Vanderbilt University \quad \IEEEauthorrefmark{3}ViSOAR LLC}
}

\maketitle
\copyrightnotice

\begin{figure*}[t]
    \centering
\includegraphics[width=\textwidth]{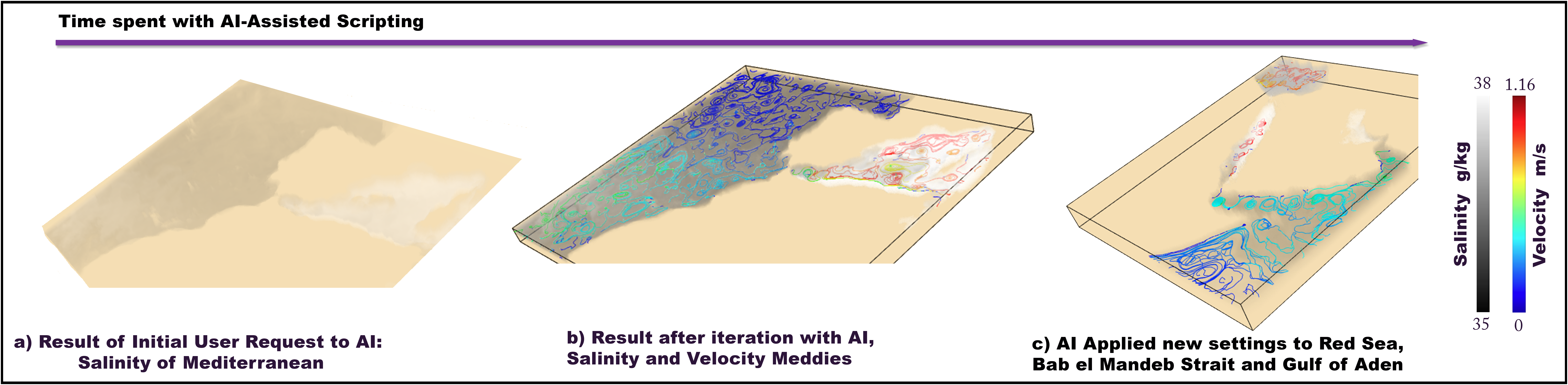}
\caption{
Our versatile framework generates a region of interest keyframe animations from petascale data using a generalized animation descriptor (GAD) file and flexible scripting. We also enable AI-assisted scripting that removes the hurdles of describing more than a region of interest to get the first image in minutes.  a) The user first asks to see the ``salinity of the Mediterranean Sea''.  After four iterations with the AI, the result in b) displays the salinity as a grayscale transfer function with streamlines to highlight the Meddies (Mediterranean eddies).  c) Further chatting with the AI applies the same visualization parameters to the Red Sea, including the Bab el Mandeb Strait and the Gulf of Aden.  Both b) and c) show regions of high salinity (white) with fast-moving currents (red) versus the oceanic regions with lower salinity (gray) and slowing currents (blue).
}
  \label{fig:teaser}
\end{figure*}

\begin{abstract}
Scientists face significant visualization challenges as time-varying datasets grow in speed and volume, often requiring specialized infrastructure and expertise to handle massive datasets. 
For example, petascale climate models generated in NASA laboratories often require a dedicated group of graphics and media experts and access to high-performance computing resources. Scientists may need to share scientific results with the community iteratively and quickly. However, the time-consuming trial-and-error process incurs significant data transfer overhead and far exceeds the time and resources allocated for typical post-analysis visualization tasks, disrupting the production workflow. 
Our paper introduces a user-friendly framework for creating 3D animations of petascale, time-varying data on a commodity workstation. 
Our proposed framework includes a streamlined animation production scripting mechanism that involves: 
(i) Generalized Animation Descriptor (GAD) with a keyframe-based adaptable abstraction for animation, 
(ii) efficient data access from cloud-hosted repositories to reduce data management overhead, 
(iii) tailored rendering system, and 
(iv) an LLM-assisted conversational interface as a scripting module to allow domain scientists with no visualization expertise to create animations of their region of interest.
We demonstrate the framework's effectiveness with two case studies: first, by generating animations in which sampling criteria are specified based on prior knowledge, and second, by generating AI-assisted animations in which sampling parameters are derived from natural-language user prompts. In all cases, we use large-scale NASA climate-oceanographic datasets that exceed 1PB in size yet achieve a fast turnaround time of 1 minute to 2 hours. Users can generate a rough draft of the animation within minutes, then seamlessly incorporate as much high-resolution data as needed for the final version. 
\end{abstract}


\section{Introduction}

Time-oriented physical phenomena are often presented as animations composed of a collection of images. Higher resolution sensors and simulations produce unimaginable data sizes with gigabytes and terabytes, with thousands of timesteps. Massive data growth results from the rapid increase in computing power, enabling closer analysis in various fields. Visualization and analysis efforts focus on distributed strategies to handle the demanding computational cost, especially for heavily parallelized algorithms. Not all domain scientists possess the supercomputer resources required for processing or storing petascale datasets for visualization. While hardware development continues to align with Moore's law, the exponential growth in data sizes outpaces advancements in computing power. The widening gap poses significant challenges for scientists and domain experts striving to create high-quality animations for petascale data.

Professional visualization institutes, like the NASA Scientific Visualization Studio, can create cinematic scientific visualizations with large datasets by employing teams of specialists and advanced software tools \cite{borkiewicz2019cinematic,borkiewicz2020introduction}. An example of this, \textit{Inside Hurricane Maria in 360°} project~\cite{borkiewicz2019cinematic} , involved four scientists, four visualizers, two media professionals, and four system support members, as well as five software platforms, including IDL, Maya, RIB, OSL, and Renderman. By transforming large-scale real-world phenomena into stunning graphics, these animations have captured millions worldwide through documentaries on large and small screens, demonstrating the power of animation in scientific communication \cite{jensen2022new}. Although domain scientists working with these datasets on commodity hardware may not need cinematic-quality animations, they still face significant barriers in computational knowledge, visualization expertise, and cross-application workflows. Although some movie-style features, such as audio narratives or a complete storyline, can be skipped in scientific sharing, data filtering, excessive application-specific programming, and visualization steps are still unavoidable. As a result, this type of project is far from available to general scientific research groups.


Our paper introduces an animation production framework for large-scale scientific datasets that operates on commodity workstations without specialized hardware. Our goal is not to replace the cinematic excellence of current large-scale approaches but to allow anyone on commodity hardware to create animations of petascale data. The contributions of our paper include:
\begin{itemize}  
    
    \item An independent language layer named Generalized Animation Descriptor (GAD) for generalized animation representation enabling highly customizable workflow and easy translation to various visualization applications.

    \item An LLM-assisted conversation-style animation scripting module that ensures user-friendly generation of time-varying animations. 
    \item Two real-world case studies with NASA's DYAMOND oceanographic datasets \cite{stevens2019dyamond, NASADYAM} on a commodity desktop machine, demonstrating the ability to create a broad range of quality-intensive scientific animations.

\end{itemize}


\section{Related Work}

In this section, we examine the role of animation as a critical tool for scientific understanding and communication, review techniques for efficient representation and rendering, and LLM-based autonomous tools in the visualization domain.

\subsection{Large-Scale Time-Varying Data Visualization}
Developing professional graphics APIs, such as OpenGL and DirectX, enabled interactive scientific data visualization on commodity hardware, as discussed in a 1994 SIGGRAPH course~\cite{hibbard1994case}. However, constrained by slower I/O, limited memory capacity, and a small budget for interactive 3D rendering, big data is a common challenge in modern scientific data visualization, especially on desktop machines~\cite{yoo2022big,Healey99,han2022narrative}. A 2010 VIS seminar~\cite{ma2011scientific} discussed scientific storytelling to better describe the findings in large-scale data.  Visualization is especially effective in communicating complex scientific concepts, encouraging collaboration with policymakers and fostering connection through public education~\cite{de2014visualization,bremer2023visualization}.

The scientific visualization community often uses widely available open-source tools for large-scale data visualization. Paraview \cite{squillacote2007paraview},
VisIt \cite{data2012visit} is a widely used open-source application that offers such functionalities. VTK \cite{schroeder2000visualizing} has served as the underlying library for many of these mainstream systems due to its comprehensiveness in object representations and file format support. Our animation production framework is designed to leverage these scientific visualization and rendering technologies as backends for different visualization needs. OpenVisus \cite{pascucci2012visus,panta2024web} offered an efficient method to stream large-scale scientific data stored in the IDX file format, enabling access to data volumes at multiple resolutions, and will be used as the underlying data loading mechanism from which we build a disk-efficient data management system aimed at generating animations for scientific analysis.

\subsection{Animation Production Frameworks for Scientific Data}
Despite the time-consuming nature of the process, animation is a compelling presentation model to convey a time-dependent scientific process. Stemming from both media editing and scientific analysis, fundamental operations in scientific data visualizations have formed animated views, such as cinematics \cite{christie2008camera,hsu2013multi}, path design \cite{liao2014storytelling,ahmed2005automatic}, and parameter transfer function adjustments \cite{jankun2001study,diaz2016adaptive}. The basic idea of stitching images is often adopted to produce flipbook videos. The most popular idea is keyframing, which records only the necessary frames and then interpolates the camera position transfer function and parameter space via linear or spline curves for smooth transitioning \cite{burtnyk1971computer,izani2003keyframe,ma2011scientific}.

\begin{figure*}[!t]
    \centering
    \includegraphics[width=0.9\linewidth]{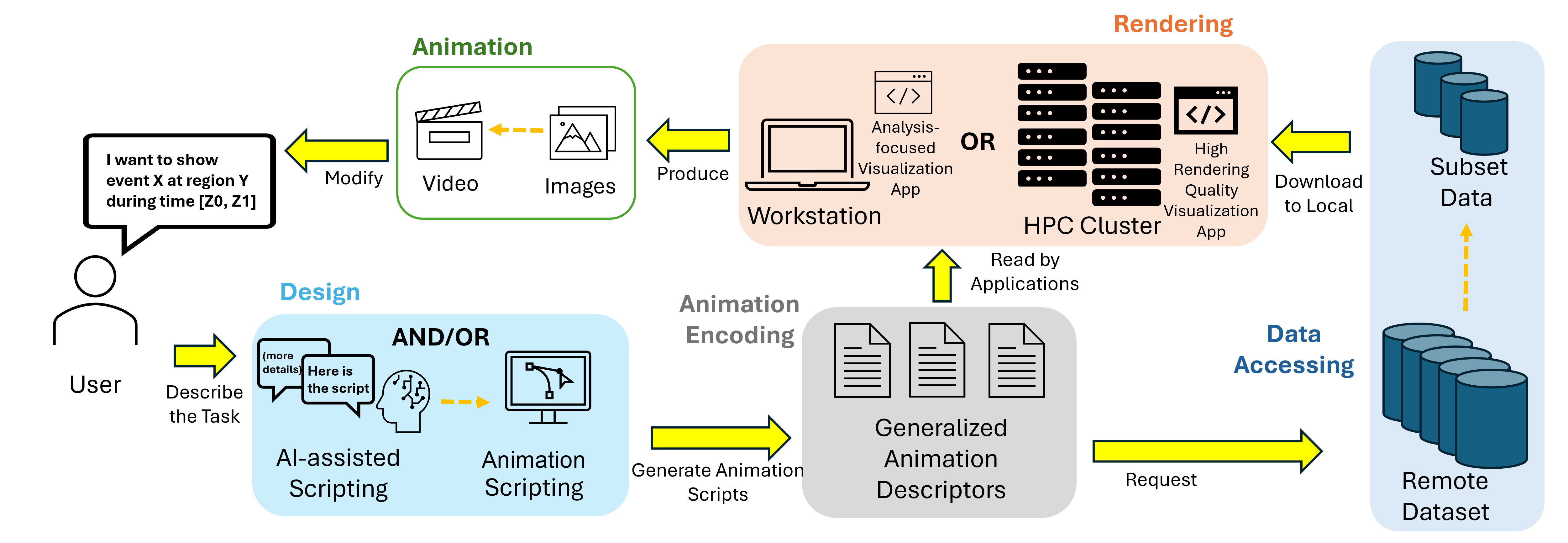}
    \caption{Our framework. Unlike traditional visualizations, where the user faces an initial big data management challenge with large-than-disk, cloud-hosted datasets, our framework starts with the event description. An optional addition of an AI-assisted scripting mechanism presents an intuitive environment to translate the conceptual design into GAD for application-independent visualization. By hiding the complexity of data management and cross-application translation, our framework simplifies the animation production cycle for the user with the illusion of directly getting a rendered video from a scripting interface with full remote dataset access.}
    \label{fig:pipeline_overview}
    \vspace{-4mm}
\end{figure*}

To align with the analysis pipeline, some abstractions were introduced, such as an event graph~\cite{Yu10}, template-based animation~\cite{akiba2009aniviz}, image-driven animation~\cite {baecker1969picture}, analysis-driven methods~\cite{nouanesengsy2014adr}, or hypercubes~\cite{beshers1988real} to incorporate domain knowledge into efficient animated visualization design. The template-based AniViz~\cite{akiba2009aniviz} system offered an abstraction for producing narrative visualization. Such domain-specific solutions were tailored to a small group of people with a narrower focus, making them difficult to extend to general data pipelines. Our paper addresses these limitations by introducing a lightweight file format that enables cross-application compatibility. Our framework can generate these descriptors through an LLM-assisted, natural-language-like conversation style, giving domain scientists greater flexibility.

\subsection{LLMs in Visualization}
The recent surge in LLMs opens the opportunity to interact in  
Natural language, offering intuitive interfaces for domain experts to query, manipulate, and generate visual outputs~\cite{kim2024phenoflowhumanllmdrivenvisual, Tian_2025, mcnutt2025accessible}. A work by Praneeth et al.~\cite{10724113} introduces a data visualization tool that leverages the power of Large Language Models (LLMs) to extract meaningful insights from vast datasets. Some tools, such as NL4DV \cite{narechania2020nl4dv}, Chat2Vis \cite{maddigan2023chat2vis}, NLI4VolVis~\cite{ai2025nli4volvis} already support generating scientific visualizations from natural language but lack the capabilities for advanced scientific analysis, such as time-series understanding of scientific data. Shusen et al. have shown that LLMs can serve as autonomous visualization agents and provide feedback for action given a visualization~\cite{ava_liu_2024}. It has also been evaluated through an information visualization literacy test~\cite{bendeck2024empirical,li2024visualization}. Due to its strong potential for visualization tasks, we find that an MLLM can be a valuable tool for generating well-structured JSON-based animation abstraction files via programmatic scripts in our framework.

\section{Our Streamlined Animation Production Framework}

We introduce a framework for animating large-scale, time-varying scientific data on commodity hardware, illustrated in Figure~\ref{fig:pipeline_overview}. In this section, we discuss the modules of our framework: 
the components of our application-independent, human-readable GAD (\Cref{sec:GAD}) scripts, how a user-determined multiresolution subset of the datasets is read and downloaded to the local disk (\Cref{dataaccess}) and rendered by the renderer backend (\Cref{render}), and the pipeline of our scripting mechanism \Cref{scripting} that generates the GAD scripts.


\subsection{Generalized Animation Descriptor} \label{sec:GAD}

\begin{figure*}[!t]
    \centering
    \includegraphics[width=0.75\textwidth]{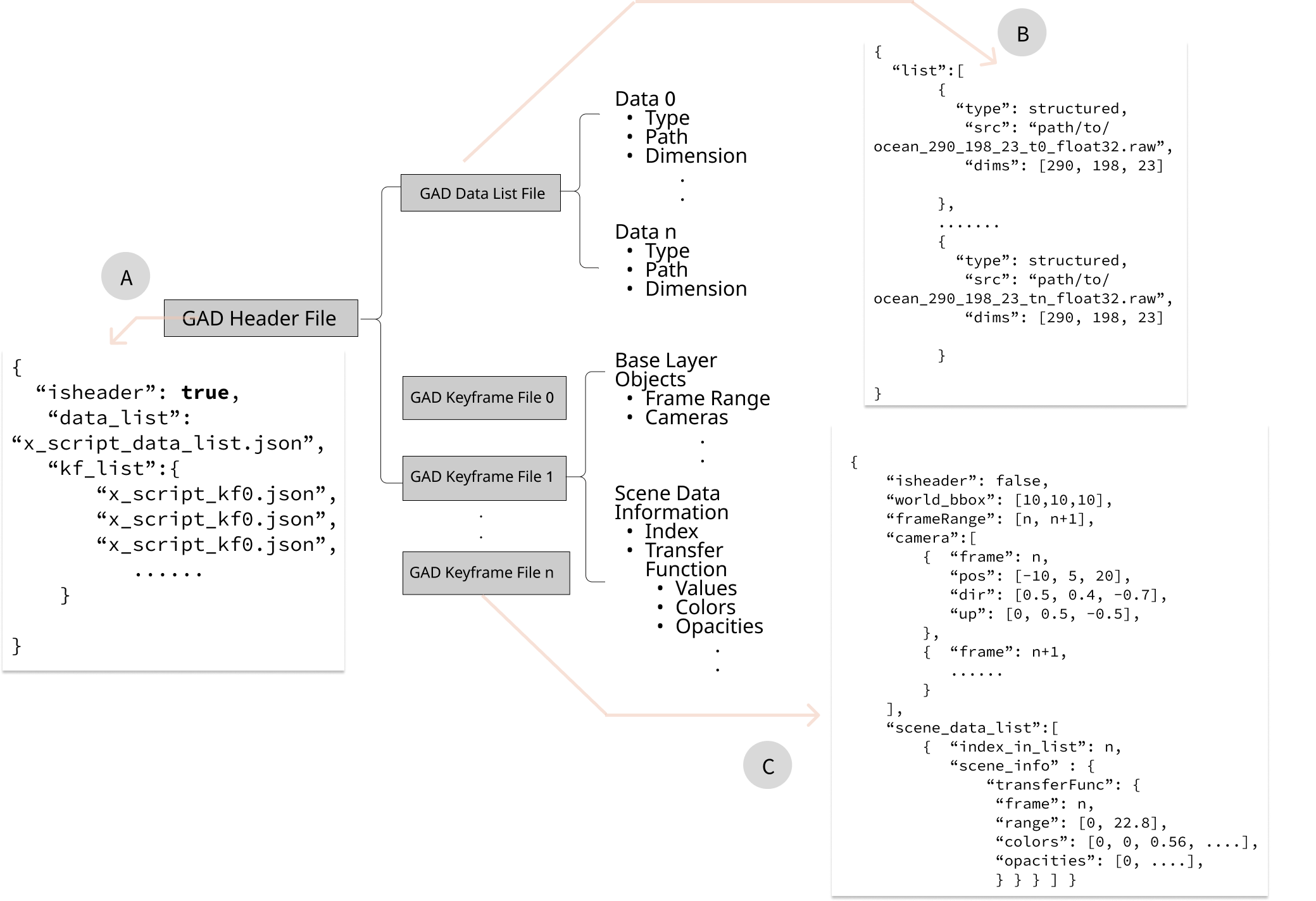}
    \caption{Our novel Generalized Animation Descriptor (GAD) file format. GAD describes an animation as a sequence of keyframes stored in an individual file. The list of data is recorded separately and accessed by keyframe files through indexing. Our modular design splits data storage and rendering information into two pieces, allowing an independent description of animation design regardless of the dataset in use.}
    \label{fig:GAD}
    \vspace{-3mm}
\end{figure*}

Our {\bf Generalized Animation Descriptor (GAD)} file format aims to represent a complex animation workflow as an abstraction layer providing a standardized representation. The GAD file format is designed to be a JSON-like keyframe system, where each keyframe contains information about static scenes and their temporal interpolation. Parameters related to a specific subset of data are passed to generate GAD files with three layers of abstraction:
(a) a header file, (b) a data list file, and (c) a series of keyframe description files as shown in Figure~\ref{fig:GAD}. GAD's hierarchical structure facilitates multi-resolution data handling, enabling users to prototype at lower resolutions before committing to resource-intensive rendering. 

The header file stores a list of keyframes and data files. The data list file stores the local data-storage paths and dimensions and the abstract data-type references as string descriptions. We currently support two data types: ``\texttt{structured}'', referring to regular gridded patterns, and ``\texttt{streamline}'', referring to a structured grid with a specific visualization technique for vector fields. However, this model is easily extendable by integrating with underlying visualization libraries while maintaining a consistent conceptual model. For example, rectilinear grid data must be manually converted to an unstructured geometry for a low-level graphics API, or they can be defined in mainstream visualization packages such as VTK using \textit{vtkRectilinearGrid} \cite{vtkVTKVtkRectilinearGrid} for ease of use and built-in optimization. The keyframe file stores all the information for the rendering tasks, including the following two categories.
\begin{itemize}
    \item {\bf Base-layer} objects, such as scene bounding box, frame range, and camera (position, direction, and up vectors) for each frame.
    \item A {\bf scene data} list containing a list of scene info, including color transfer function values and opacity mappings for volume rendering.
\end{itemize}
Each attribute can be interpolated between keyframes, allowing for smooth transitions in camera movement and visualization properties. GAD's piecewise attribute definition allows animations to be designed at varying levels of detail, with the flexibility to fine-tune specific keyframes without affecting the overall animation.

One key strength of the GAD script format is its application-independent design, which frees animation workflows from the constraints of specific visualization tools. The framework only stores pointers to raw data files in the GAD Header File (\Cref{fig:GAD}(a)), metadata (\Cref{fig:GAD}(b)), and animation descriptions (\Cref{fig:GAD}(c)), allowing the target application to handle the underlying data reading and interpretation.

\subsection{Accessing Data from Remote Storage}

\label{dataaccess}
Accessing petascale data without running out of RAM or disk space on commodity hardware for 3D animation generation is not a simple task \cite{cox1997managing}. Based on the NSDF data fabric abstraction layer by Panta et al. \cite{panta2024web}, we provide user-friendly querying of scientific information from datasets while hiding the complexities of file systems and cloud services. The NSDF framework converts high-performance computing (HPC) data to cloud data in a visualization-ready IDX format~\cite{Openvisus_git} and manages data caching. Their framework proved ideal for researchers and scientists to manage petascale data. Hosted on the Open Science Data Federation~\cite{osdf}, the high-resolution atmospheric and oceanic simulation datasets are available for public access \cite{panta2024web}. OpenVisus API \cite{pascucci2012visus} calls are needed to access data in this framework. Our GAD script has the appropriate parameters saved for the subset of data the user requests. Along with the GAD script and specific downsampling queries, our Python layer manages data acquisition and downloads the subset of datasets in a local workstation.

\subsection{Data Rendering and Visualization}
\label{render}
Although the GAD script format can be used in any rendering application, we leverage the existing OSPRay \cite{ospray} and VTK-based \cite{vtk} offline rendering mechanisms to demonstrate its use and develop minimal extensions to support the GAD format. We choose OSPRay for its high-performance CPU-based ray tracing engine, memory efficiency, and rendering performance. We implement an \texttt{AnimationHandler} class that drives OSPRay and VTK rendering engines and passes the GAD files as JSON through Python bindings to OSPRay via \texttt{vistool\_{py}\_{osp}} and VTK via \texttt{vistool\_{py}\_{vtk}}. The bindings enable the same GAD script to produce consistent visualizations across different rendering backends. In the rendering backend, we write a custom \texttt{loadKF()} method to read the keyframe data from GAD files. We use \texttt{OpenGL} and \texttt{GLFW} to visualize the rendered frames as image files generated by OSPRay volume rendering tasks. Standard VTK libraries handle the volume rendering and visualization of VTK's rendered outputs. We configured both to use similar camera and background settings, acknowledging that the rendering is different between the two. 

Our implementation processes one GAD keyframe file at a time, loading the specific data needed for that keyframe, rendering it, and saving the resulting image before moving on to the next. Our approach reuses the same memory buffers for each frame (e.g., the \texttt{vtkImageData} object in the VTK implementation), achieving better memory efficiency than loading all animation data at once.

\subsection{Animation Scripting}\label{scripting}

Animation scripting plays the primary role in generating GAD scripts. Our scripting interface provides an easy-to-navigate layer to create 3D animations from a large cloud-hosted dataset without a steep learning curve or visualization expertise. Our animation scripting has two methods: (a) basic Python scripting and (b) a multimodal large language model (MLLM) assisted conversation-style scripting modules. The former suits those who know the exact parameter inputs that work well for a particular visualization task. The latter is for a quick start and allows explorations through conversation with AI. The scripting mechanisms record the region of interest in our GAD files and download the requested data subset in preparation for visualization.

\subsubsection{Basic Scripting with Python}\label{text}
\begin{figure}[!t]
\centering
    \includegraphics[height=.5\textwidth]{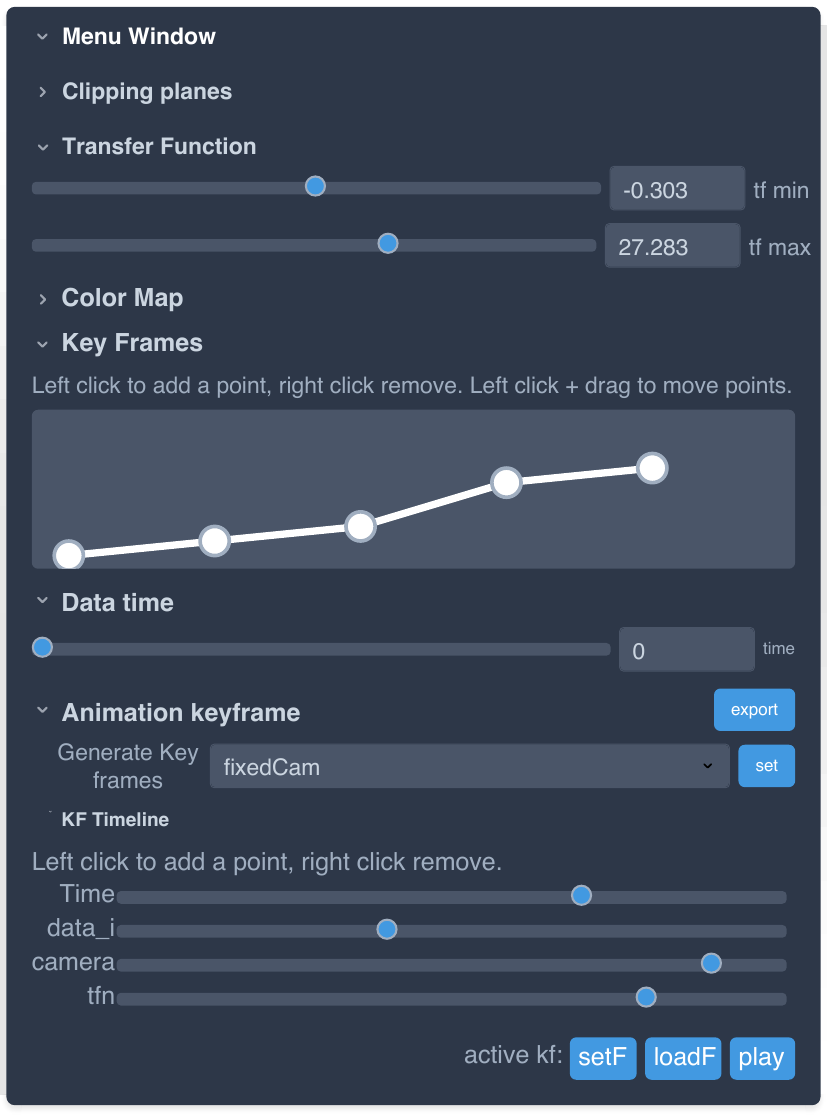}
    \caption{Our simple interactive viewer provides a quick interface for users to create and iteratively modify animations with just a few settings. 
    }
    \label{fig:viewer}
\end{figure}
Our basic Python scripting module includes two modes: a command-line interface and an interactive viewer. In the first mode, the user can input the values of the x, y, and z coordinate parameters, timesteps, camera positions, and data quality by typing them at the command line. The user has to know the coordinates of their region of interest. They also need to understand the complexities of specifying a camera in 3D. To facilitate that, the user can use the second method, which has an initial prototype of an interactive viewer to assist with choosing the region of interest. Our viewer is implemented in C++ with a Python layer that directly streams the corresponding data into memory. We use \texttt{ImGui}~\cite{knoblauch2017imgui}
for the UI implementation to adjust scene elements such as transfer functions, clipping planes, and animation controls by adding keyframes (\Cref{fig:viewer}). When the user clicks the ``Export'' button, the system communicates with the Python module, reads the data, and downloads the GAD scripts with raw data. The viewer-based method is more manageable for some users than the text-based method; however, it still requires understanding terms like transfer functions, clipping, and animation keyframes.

\subsubsection{LLM-Assisted Scripting}

Our LLM-assisted scripting is implemented as a conversational ChatGPT-like user interaction, similar to the web-based application familiar to the general public. We use the GPT-4o model to build our rule-based AI-assisted~\cite{servantez2024chain} scripting framework. The key principle of our scripting mechanism is to leverage AI's assessment and visual understanding capabilities in a context-aware manner. Our scripting mechanism comprises four parts: context building, action planning, animation evaluation, and memory.

\paragraph{Context-building}
As part of context building, we summarize the dataset during initialization and provide a parameter template consistent with the GAD files. In the intermediate step, a function call is made to set the parameters when the user asks for a specific animation. Inside the function, as context, the system provides the MLLM with predefined examples of successful parameter sets for known events related to our example dataset. Throughout the conversation, the MLLM preserves all conversation history and passes it as part of the prompt, enabling the AI to learn the user's request context over time. 

\paragraph{Action planning}
We show users a command-line interface with options 1-4 to choose an example phenomenon, or option 0 to enter a custom description of what they want to animate. In the case of options 1-4, the script looks for example parameters and, if present, the pre-generated GAD scripts, and further, the render backend works in line. In the case of option 0, when users describe a desired visualization in natural language, the MLLM uses these examples and all the contexts passed to it to translate the description into specific parameter values ($x_range$, $y_range$, time steps, etc.). The script then sets the parameters in appropriate function calls, generates corresponding GAD scripts, and renders the animation. 

\paragraph{Evaluating animations}
We enable AI-empowered animation feedback by returning the rendering results to the MLLM. Based on all the contexts provided to the MLLM, it assesses the rendered frames and suggests parameter adjustments to improve visualization. Then it asks the user whether they want to animate using the suggested parameters or guide MLLM in making modifications. Upon the user's input, it returns adjusted parameters and continues this loop until the user exits the interface.
 
\paragraph{Memory}
All these components need is the memory of the previous actions or the visualization outputs it observed before. To make our scripting more time-efficient, each animation is assigned a unique identifier that encodes the selected spatial bounds, timestep range, data resolution, field, and streamline settings. If the parameters for the region of interest match a locally saved animation, the system reuses the stored GAD scripts and renders the animation directly. In this way, redundant data access, load, and GAD generation are prevented. This represents a step towards a fully automated visualization framework with a highly human-interpretable user interface.

All four components of the AI-assisted scripting mechanism work together to create a user-friendly environment that allows non-visualization experts to generate animations.

\section{Overview of Cloud-Hosted Dataset}
\label{dataset}
\begin{figure}[htbp]
    \centering
    \includegraphics[height=0.27\textwidth]{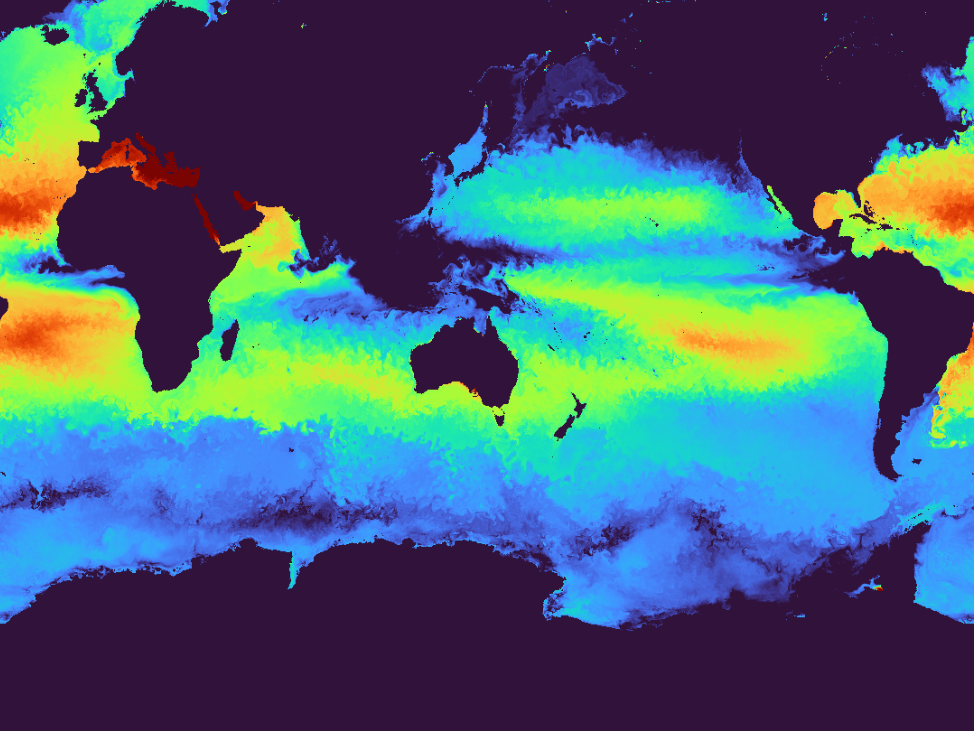}\hfill
    \includegraphics[height=0.23\textwidth]{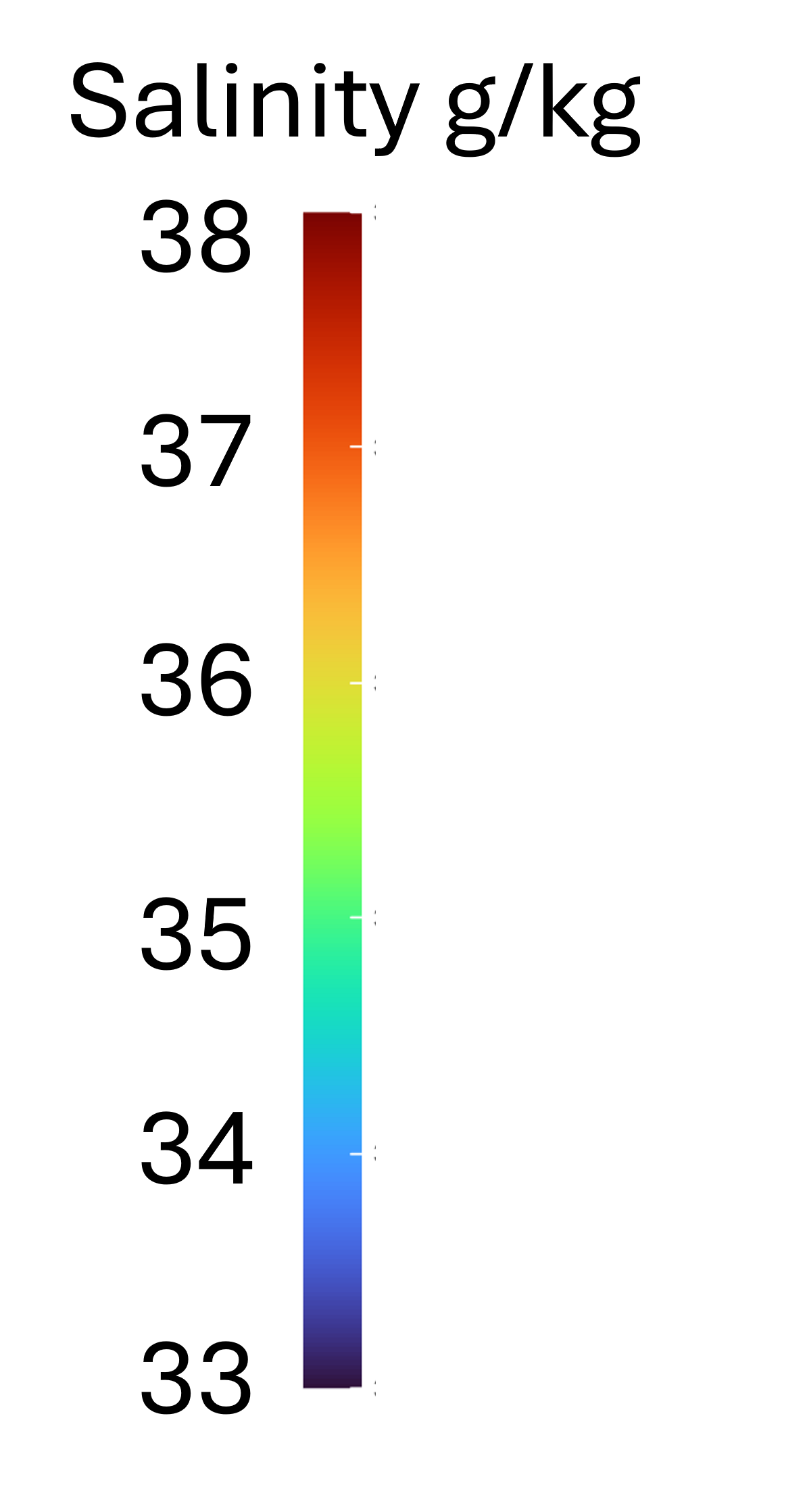}
    \caption{An ocean surface (Z=0) salinity map, ranging from 33 to 38 g/kg. This visualization represents one of 10,269 timesteps from a 1PB dataset, each 20GB in size. The complete dataset includes five 3D scalar fields: temperature, salinity, and velocity components east-west, south-north, and vertical.}
    \label{fig:data_sizes}
\end{figure}

For the results in this paper, we use the oceanographic simulation of the NASA 1.8 PB DYAMOND dataset \cite{stevens2019dyamond, NASADYAM}, shown in Figure~\ref{fig:data_sizes}. This model was run for over 10000 hourly timesteps covering over 14 simulation months starting from  January 2020 involving five 3D fields, including temperature, salinity, three axis-aligned velocities, and more than ten 2D fields. ``Dynamics of the Atmospheric General Circulation Modeled On Non-hydrostatic Domains” or DYAMOND is part of the Coupled Ocean-Atmosphere Simulation (COAS) run at \emph{NASA Advanced Supercomputing (NAS)}. These datasets are hosted on the cloud and have been publicly available through NASA's data portal \cite{dyamonddata,nasaEstimatingCirculation, sciviscontest2026}. The DYAMOND dataset well represents the big data visualization challenge our framework addresses. Identifying features of interest within this large data set is often complicated by involving multiple intermediate steps from various applications. Our framework facilitates interactive exploration of these massive datasets by decoupling the animation process from data management, allowing domain scientists to focus on discovering important oceanographic phenomena rather than struggling with visualization technicalities.

\section{Experiments}

We demonstrate our framework's effectiveness through two case studies using NASA's DYAMOND LLC2160 ocean dataset as introduced in \Cref{dataset}. Our experiments show how our approach enables visualization of petascale data on commodity hardware with fast turnaround times, ranging from one minute to two hours, depending on the complexity. We present the workflow of our framework to obtain a fixed-camera 3D animation similar to existing pre-recorded videos targeting the oceanographic phenomenon important to the domain scientists of the dataset. Our use cases are run on a commodity workstation with a 12th Gen Intel(R) Core(TM) i7-14700 CPU with 128GB of memory and a Quadro RTX 5000 GPU. Images are rendered at a screen resolution of 2048x2048 in all case studies.

\begin{figure*}[!t]
    \centering
\includegraphics[width=0.85\textwidth]{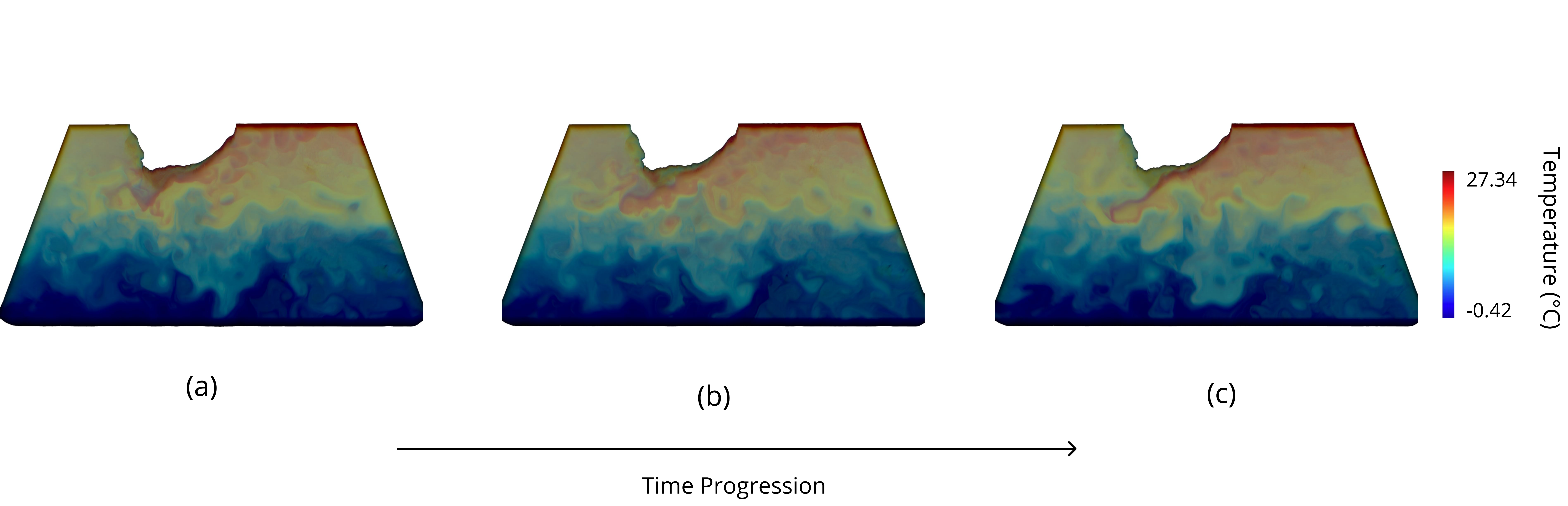}
    \caption{A user produces an animation of the ocean temperature in the Agulhas retroflection (turns back to itself) over time from January 20, 2020, to April 19, 2020. The eddies form due to retroflection and create circular structures called ``Agulhas Rings". The images show the eddies formed in yellow-red regions, these eddies spinning off the main current and traveling into the South Atlantic: (a) at timestep 1, a line of warm current starting retroflection, creating some eddies in the warmest ocean surface, (b) at timestep 24 the current rotated back with ring formation with some visible eddies in merging points with cooler currents in the Southern Ocean (c) at timestep 90 the current started rotating back again. This animation was created using our framework in 30 minutes for GAD script generation and data download, plus 12 minutes for rendering.}
    \label{fig:case1o1}
\end{figure*}

\subsection{Case Study 1: Visualizing Oceanographic Phenomena with GAD generation}

We targeted the Agulhas Ring for Case Study 1. Formed by the confluence of warm Mozambique and East Madagascar currents, the current travels southward, carrying heat and momentum from the Indian Ocean \cite{Lutjeharms2006}. 
However, as the Agulhas current loops back to itself, an eddy is formed, creating circular structures called ``Agulhas Rings" that play a significant role in ocean dynamics and can impact regional climate patterns.

\begin{figure*}[!t]
    \centering
\includegraphics[width=0.9\textwidth]{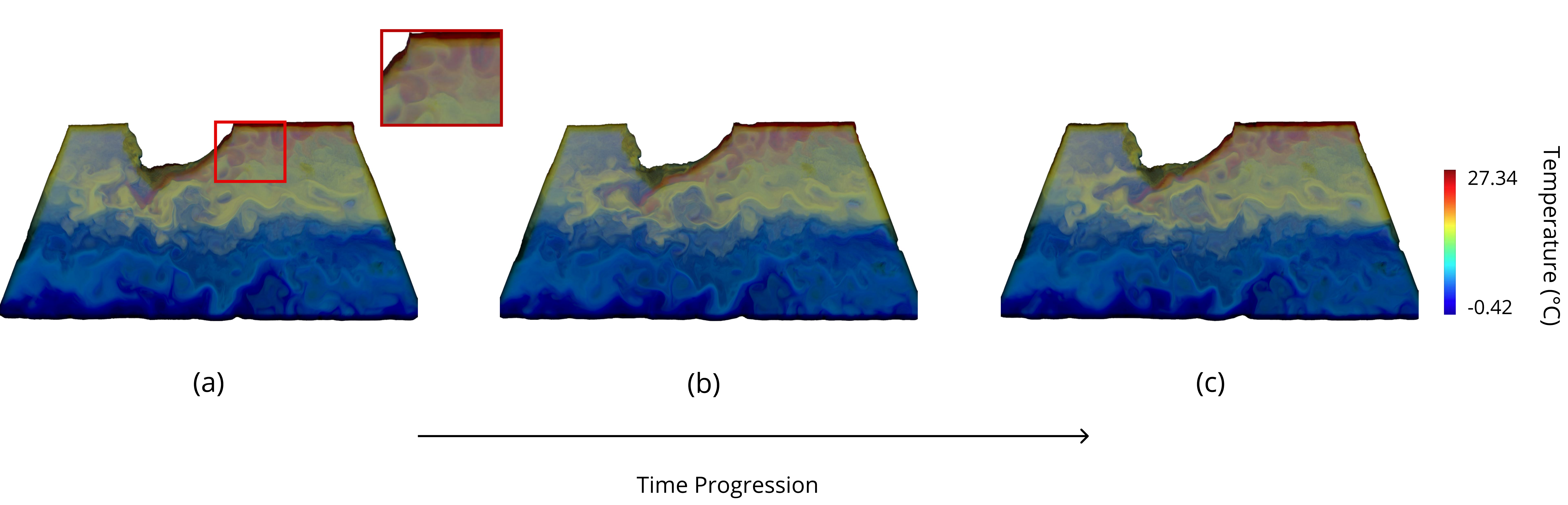}
    \caption{The user improves upon the animation in Figure~\ref{fig:case1o1} with our interactive viewer tool to include opacities. By reducing opacity in cooler regions while maintaining the visibility of warmer structures, the eddies in the Agulhas current become more prominent, especially in warmer regions. At (a) timestep 1, a line of warm water starting retroflection, showing better eddy structure in the warmest ocean surface, (b) at timestep 5, and (c) timestep 10, the retroreflections are still forming while eddies in warmer ocean surface are more prominent visibility than Figure~\ref{fig:case1o1}.  }.
    \label{fig:case1o2}
\end{figure*}

With the interactive viewer, we first identify the region of interest for the Agulhas area and pass the parameters to the Python code through the command line. We obtain 90 timesteps of full-resolution data (quality = 0), starting from timestep 0 and spaced 24 hours apart, to generate a rough animation overview for a 90-day duration. For the DYAMOND dataset, this means 2020/1/20 to 2020/4/19. In about 30 minutes, the script could generate GAD scripts and download datasets from our region of interest. We feed the GAD Header file to our rendering backend, discussed in \Cref{render}, and it takes approximately 12 minutes to render all the frames. After rendering, we visualize the animation as a GIF file to see the Ring formations; please see the supplementary materials accompanying this paper. \Cref{fig:case1o1} highlights three frames from our animation, 1st, 24th and 90th frames generated by GAD scripts.

Although the ring formation is distinguishable from the temperature map in the above images, we further refined the visualization to get a clearer ring formation view near the warmer temperature region. Using the interactive viewer, we loaded 10 representative timesteps and adjusted the transfer function for better clarity. After tuning the opacity and color mappings to achieve optimal visualization of the oceanic structures, we exported keyframes for each timestep. This process automatically generated GAD files containing 10 keyframes with precisely defined interpolated opacity values. The resulting GAD file serves as a comprehensive, self-contained description to reproduce the animation. \Cref{fig:case1o2} shows the results of these steps, and the supplementary material contains the animation.


Due to its complexity, creating 3D animations of this event has traditionally been time-consuming and daunting for most researchers. However, we demonstrate that this can now be accomplished with minimal programming knowledge and on standard workstations in a fast turnaround time.

\subsection{Case Study 2: LLM-Assisted Exploration of Multi-Variable Oceanographic Data}

In this case study, we demonstrate how our AI-assisted scripting pipeline can help generate visualizations, compare them, and suggest visualizations with multivariate fields to better understand data with a fast turnaround time. 
Case Study 2 focuses on the intriguing behaviors of the Mediterranean and Red Seas. 

\begin{figure*}[!t]
    \centering
    \includegraphics[width=0.98\textwidth]{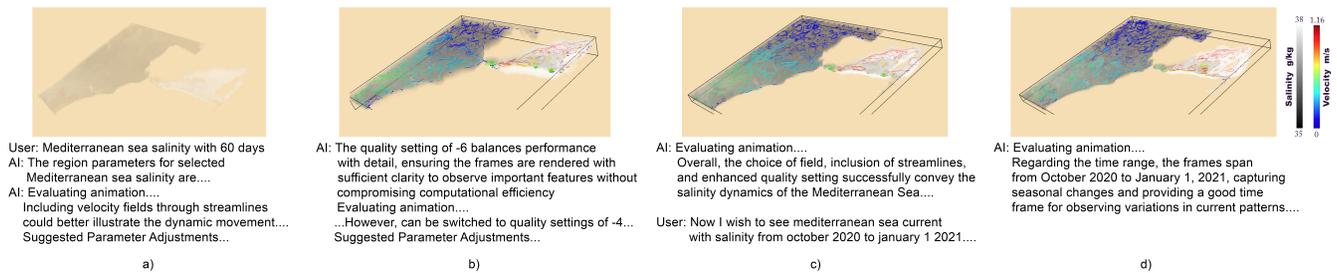}
    \caption{Portion of a user's session with AI-assisted scripting to display the salinity of the Mediterranean Sea. The first region-of-interest request results in a) a black-and-white volume of the salinity field. As the user chats with the AI more, the AI suggests visualizing both the salinity and the velocity fields as streamlines, shown in b), where streamlines are traced from the combined u, v, and w velocity vector fields. Image c) shows refined parameters.  Then, the user asks to apply the current settings to a different time frame, as shown in d). This region of interest showcases the dynamics near the Straits of Gibraltar, where fresh Atlantic water flows into the salty Mediterranean Sea at the surface, and denser water flows out at the bottom, forming rotating, lens-shaped Meddies\cite{meddy}.}
    \label{fig:case2}
     \vspace{-3mm}
\end{figure*}


\paragraph{Mediterranean Sea Salinity}
The Mediterranean Sea represents a classic example of an evaporative basin with unique water circulation patterns. At the Strait of Gibraltar, fresher Atlantic water flows into the Mediterranean at the surface. In contrast, saltier, denser Mediterranean water flows outward at depth, eventually forming rotating, lens-shaped bodies called Meddies \cite{meddy}.


To explore this region, we launch our animation scripting module and choose option 0 to create a custom description in natural language at the AI prompt. We input with ``Mediterranean sea salinity with 60 days''. The AI chooses a timeframe of 60 days, 24 hours apart, starting from 2020/01/20 to 2020/12/12 with 1/256 of the original resolution, and adds no streamlines at first. It took about 9 seconds to download the data and 1 minute to render. Five resulting frames with salinity volume are then sent to the MLLM for evaluation and adjustment suggestions. The AI suggested slightly modifying the x, y, and z coordinates, increasing the resolution to 1/64 of the original resolution, and switching to visualizing salinity with streamlines. This iteration took 15 minutes to download the new, higher-resolution data and approximately 1 minute to render. In the next evaluation, the frames are generated with 1/16 of the original resolution.  The user was satisfied after four iterations in this evaluation-action-memory loop. Then, the user explored a different date range, asking the AI to create the same animation from October 2020 to January 2021. The final animation ended up with a resolution of 1/16 of the original data downloaded. \Cref{fig:case2} provides a sampling of this AI-assisted scripting loop, which proves that iterative exploration can help domain users with an initial screening of a large-scale dataset. The full dialogue of this session is provided in the supplementary materials.

\paragraph{Red Sea Salinity}
To test our system's flexibility with regions beyond the example regions (Agulhas Ring and Mediterranean Sea) set during context-building, we ask the AI to animate the eddies in the Red Sea with the description: ``I want to see Red Sea Currents with Salinity''. The Red Sea is a sea inlet in the Indian Ocean known for its high salinity due to its high evaporation and low precipitation. Eddies in the Red Sea, particularly in the central and northern regions, play a crucial role in transporting salt. Our goal is to animate the eddies over time. Without any prior examples specific to this region, the AI initially struggled to suggest appropriate coordinates for the Red Sea and captured somewhere near southeastern Africa in the Indian Ocean. After some guidance like ``please capture a broader range for x and a different range at the upper regions", the AI identified coordinates encompassing the Red Sea with some part of the Gulf of Aden, capturing the critical Bab-el-Mandeb strait where these water bodies connect from January 20, 2020, to January 30, 202, 24 hours apart. Figure~\ref{fig:teaser}c shows one of the rendered frames with distinctive eddy formations in this strait and the center of the Red Sea, highlighting the circular currents.


This case study reflect the effectiveness of AI-assisted mechanisms to help domain scientists set parameters and meta information for animation production without rigorous visualization expertise. Users simply need to select the dataset they are interested in, and then they can have a natural-language conversation with the AI to receive focus-based guidance.

\section{Limitations and Further Research}

We have shown that our Generalized Animation Descriptor (GAD) can be generated with LLM-assisted scripting and integrated with traditional visualization packages. One limitation is, the MLLM API has inherent randomness in its inference process, resulting in slightly different outputs in each iteration.In the future, this conversational interface may suggest suboptimal parameters that require manual refinement by users. Moreover, fine-tuning the MLLM for the 3D animation generation pipeline will boost the framework's capabilities. Our preliminary framework takes a big step forward, reducing the learning curve for domain scientists and researchers in limited-resource environments. 

We envision adding PyVista and ANARI in our framework. PyVista provide a more ready-to-use scenario and ANARI shares similarities with GAD in providing a standardized interface. Since each tool has its own use case, the user experience could be improved by providing more backend suggestions tailored to the animation design. In the future, rather than relying on the general MLLM model. By defining a robust evaluation metric for the quality of scientific animations, the next objective could be to create a fully automated, self-improving animation generation system that eliminates the need for iterative manual modifications.

\section{Conclusion}

Large-scale scientific datasets present significant visualization challenges for domain scientists, requiring specialized expertise and infrastructure that often exceed available resources. Our comprehensive framework enables efficient handling of petascale, cloud-hosted datasets through progressive refinement workflows that begin with rapid low-resolution prototyping and advance to high-quality animations. The GAD abstraction provides a platform-agnostic method for animation generation, allowing seamless integration with existing visualization tools (OSPRay, VTK) while hiding underlying technical complexities from users. The natural language interface further reduces barriers by translating domain-specific descriptions into technical parameters, eliminating the need for scientists to understand coordinate systems or visualization-specific settings. Through two real-world case studies using NASA's petascale DYAMOND oceanographic dataset, we demonstrate that domain scientists can produce high-quality 3D scientific animations on commodity hardware without requiring specialized visualization expertise. 


Our approach enables researchers to focus on scientific discovery and analysis rather than on the technicalities of visualization, democratizing access to petascale data visualization capabilities previously limited to specialized visualization teams with significant computational resources.

\section*{Acknowledgment}
 
This work was funded by NSF OAC award 2138811, NSF CI CoE Award 2127548, NSF OISE Award 2330582, the NASA AMES cooperative agreements 80NSSC23M0013, and NASA JPL Subcontract No. 1685389.In part, the work was done under the auspices of the DoE by LLNL under contract DE-AC52-07NA27344 (LDRD project SI-20-001) and DOE SBIR Phase II \#DE-SC0017152. Open Science Data Federation(OSDF), and ViSOAR LLC (\texttt{https://visoar.com/}) provided cloud storage to host the data. 
\bibliographystyle{ieeetr}

\bibliography{template}


\end{document}